%% file: main.tex
\pdfoutput=1

\documentclass[11pt]{article}
\usepackage{acl}
\usepackage{multirow}
\usepackage{times}
\usepackage{latexsym}
\usepackage[T1]{fontenc}

\usepackage[utf8]{inputenc}

\usepackage{microtype}

\definecolor{mint}{rgb}{0.24, 0.71, 0.54}
\usepackage{soul}
\usepackage{graphicx}
\usepackage{hyperref}
\usepackage{caption}

\title{SOCCER: An Information-Sparse Discourse State Tracking Collection\\in the Sports Commentary Domain}

\author{Ruochen Zhang \and Carsten Eickhoff\\
       Brown University \\
        \texttt{\{ruochen\_zhang, carsten\}@brown.edu} \\}

\begin{document}
\maketitle
\begin{abstract}
In the pursuit of natural language understanding, there has been a long standing interest in tracking state changes throughout narratives. Impressive progress has been made in modeling the state of transaction-centric dialogues and procedural texts. However, this problem has been less intensively studied in the realm of general discourse where ground truth descriptions of states may be loosely defined and state changes are less densely distributed over utterances. This paper proposes to turn to simplified, fully observable systems that show some of these properties: Sports events. We curated 2,263 soccer matches including time-stamped natural language commentary accompanied by discrete events such as a team scoring goals, switching players or being penalized with cards. We propose a new task formulation where, given paragraphs of commentary of a game at different timestamps, the system is asked to recognize the occurrence of in-game events. This domain allows for rich descriptions of state while avoiding the complexities of many other real-world settings. As an initial point of performance measurement, we include two baseline methods from the perspectives of sentence classification with temporal dependence and current state-of-the-art generative model, respectively, and demonstrate that even sophisticated existing methods struggle on the state tracking task when the definition of state broadens or non-event chatter becomes prevalent. 
\end{abstract}
\begin{figure*}[t]
\centering
    \includegraphics[width=1\linewidth]{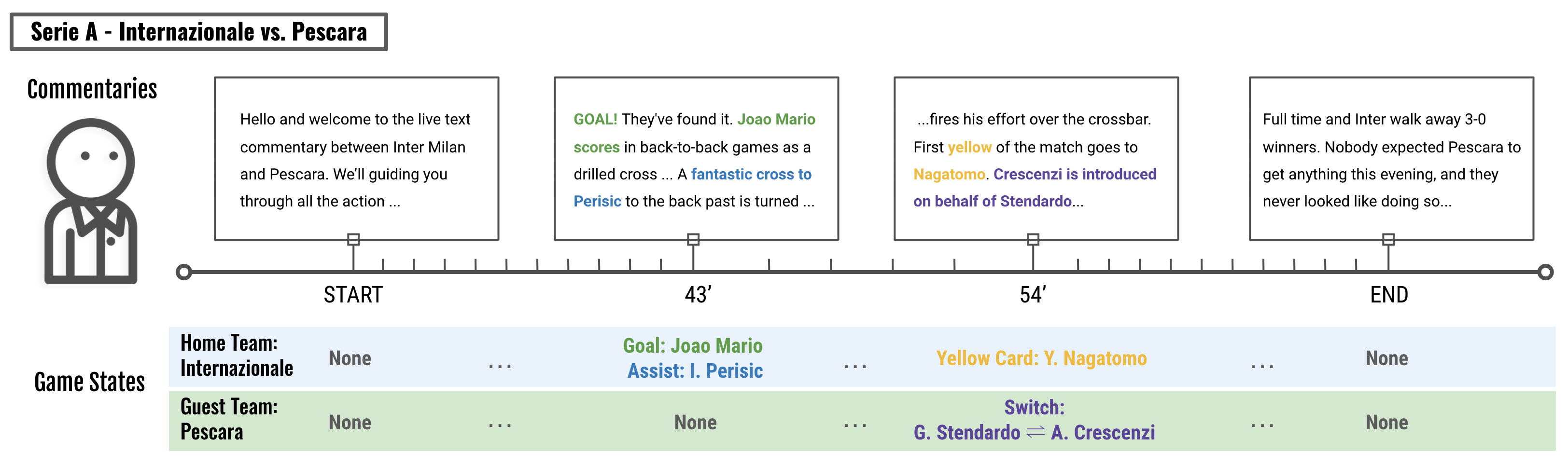}
    \caption{\label{fig:overview} An overview of the state tracking task in sports commentary.}
\end{figure*}

\section{Introduction}

State tracking, the task of maintaining explicit representations of user requests and agent responses, has long been a key component of dialogue systems \citep{williams-etal-2013-dialog, henderson-etal-2014-second, henderson2014third, kim2016fifth}. The same challenge arises during reading comprehension of procedural texts (recipes, how-to guides, etc.) where systems focus on predicting changes of object attributes at the entity-level (a car window may transition from foggy to clear) \citep{dalvi-etal-2018-tracking, tandon2020dataset}. However, both of these state tracking variants rely on transaction-based or turn-based data such as transactional dialogues or procedure descriptions that are information-dense. Few works have studied state tracking tasks where state changes occur infrequently while a large proportion of messages are ``chatter''.

As an alternative to altogether unrestricted state tracking---a task that is daunting due to the complexity of even describing ground-truth states in a discrete manner---we resort to a simpler and more self-contained setting: sports competitions. Given the stream of natural language utterances with which a commentator describes the events in a real-world setting (here a sports competition), an ideal natural language understanding system would maintain and reason over a coherent and accurate representation of the match based on how the commentator described it. This representation can, in turn, be used for downstream tasks such as inference or language generation. Sports matches provide an ideal test bed for state tracking due to their self-contained, fully observable nature and their inherent interpretability in the form of the temporal evolution of scores. However, existing sports-related commentary collections such as described by \citet{aull2013fighting} and \citet{merullo-etal-2019-investigating} do not provide such within-match temporal information.

To this end, we collect temporally-aligned commentaries and live scores of soccer matches along with other meta information from the website \href{https://www.goal.com/en-us}{goal.com} and compile the dataset \texttt{SOCCER}. To the best of our knowledge, \texttt{SOCCER} is the first temporally-aligned collection of sports match commentary and state. It contains over 2,200 matches from tournaments such as the UEFA Champions League or the UK Premier League between 2016 and 2020. Across these matches, there are over 135,000 individual comments and approximately 31,000 events. A simplified example is shown in Figure~\ref{fig:overview}.

To demonstrate the potential of state tracking for open-domain discourse, we use the proposed dataset to investigate to what degree state-of-the-art systems are able to track the progression of events described in the commentary. This overview includes two model classes: classification models that treat match events as different class labels, and generative language models such as GPT-2 \citep{radford2019language} that model context and events in a causal manner. Our experiments show that both methods do not perform well on \texttt{SOCCER} and only slightly outperform distributional heuristics, leaving considerable room for improvement. 

The novel contributions of this paper are three-fold: (1) we propose a new task of tracking event occurrences via state changes, (2) we create \texttt{SOCCER}, a general discourse state tracking dataset that contains temporally-aligned human-composed commentary and in-game events, serving as the training and evaluation dataset for this task, and (3) we provide two intuitive baselines demonstrating the difficulty of this task and presenting exciting opportunities for future research. 

\section{Related Work}

\textbf{Dialogue State Tracking (DST).} Current DST collections and benchmarks tend to rely on transaction-centric dialogues with predefined domain-specific ontologies and slot-value pairs. Prominent examples include the DSTC2 \citep{henderson-etal-2014-second} and MultiWOZ datasets \citep{budzianowski2018multiwoz}. Consequently, previous work focuses on picklist-based approaches \citep{mrksic-etal-2017-neural, perez-liu-2017-dialog, zhong-etal-2018-global, ramadan-etal-2018-large,gao-etal-2019-dialog} to formulate state tracking as a series of classification tasks over candidate-value lists. A major difference between \texttt{SOCCER} and other DST datasets lies in its information density. As dialogues in DST are usually short conversations with direct transactional objectives such as booking hotels or reserving restaurant tables, frequent state changes are required to be captured within limited turns of the conversation. In sports commentary, on the contrary, in-game events occur at a comparatively low frequency and a considerable proportion of commentator utterances may not be related to any changes in the game state.

\textbf{State Tracking in Procedural Text.} State tracking in procedural text understanding focuses on the task of tracking changes in entity attributes \citep{tandon2020dataset}. A variety of procedural progresses have been proposed such as tracking entity presence and location in scientific processes \citep{dalvi-etal-2018-tracking}, ingredients in cooking recipes \citep{bosselut2017simulating}, and character motivation and emotional reaction in simple stories \citep{rashkin2018modeling}. Yet, similar to DST settings, these highly specific tasks depend on small fixed ontologies covering limited ranges of entities and states. Another more recent dataset \citep{tandon2020dataset} turns to an open-vocabulary setting when defining entity attributes. But since the dataset is comprised of how-to guides from WikiHow.com, the task still sees a high density of state changes per natural language instruction. 

\textbf{Information Density} 
The concept of Information Density has been mainly used in the Uniform Information Density (UID) theory \citep{jaeger2010redundancy} to measure the amount of information per unit comprising an utterance. \citet{levy2007speakers} demonstrated that speakers tend to maximize the uniformity of information via syntactic reduction. The notion of information density in our paper, however, focuses on quantifying the frequency of event occurrences on the corpus level instead of understanding syntactic choices on the utterance level.

\textbf{Sports Event Datasets and Tasks.} Commentary in the sports domain has been collected to study a variety of problems such as racial bias in football game reporting \citep{merullo-etal-2019-investigating} and gender construction in NBA/WNBA coverage \citep{aull2013fighting}. However, these datasets do not provide any information on the temporal alignment between commentary and events. Another similar dataset, \texttt{BALLGAME} \citep{keshet-etal-2011-ballgame} is comprised of baseball commentary with annotated events and timestamps, but it contains less than 20 games and the annotation is unavailable online. Some work focuses on sports-related inference of player performance metrics \citep{oved2019predicting} or game outcomes \cite{velichkov-etal-2019-deep} that predict full-time results based on signals from pre-game player interviews. However, no in-game sequential contexts are provided in these datasets. Most similar to our work, \citet{bhagat2018towards} collected in-game commentaries for soccer player analytics, but their approach is restricted by classical machine learning methods and ignores the effect of information sparsity within the dataset.
\begin{figure}[t]
\centering
    \includegraphics[width=1\linewidth]{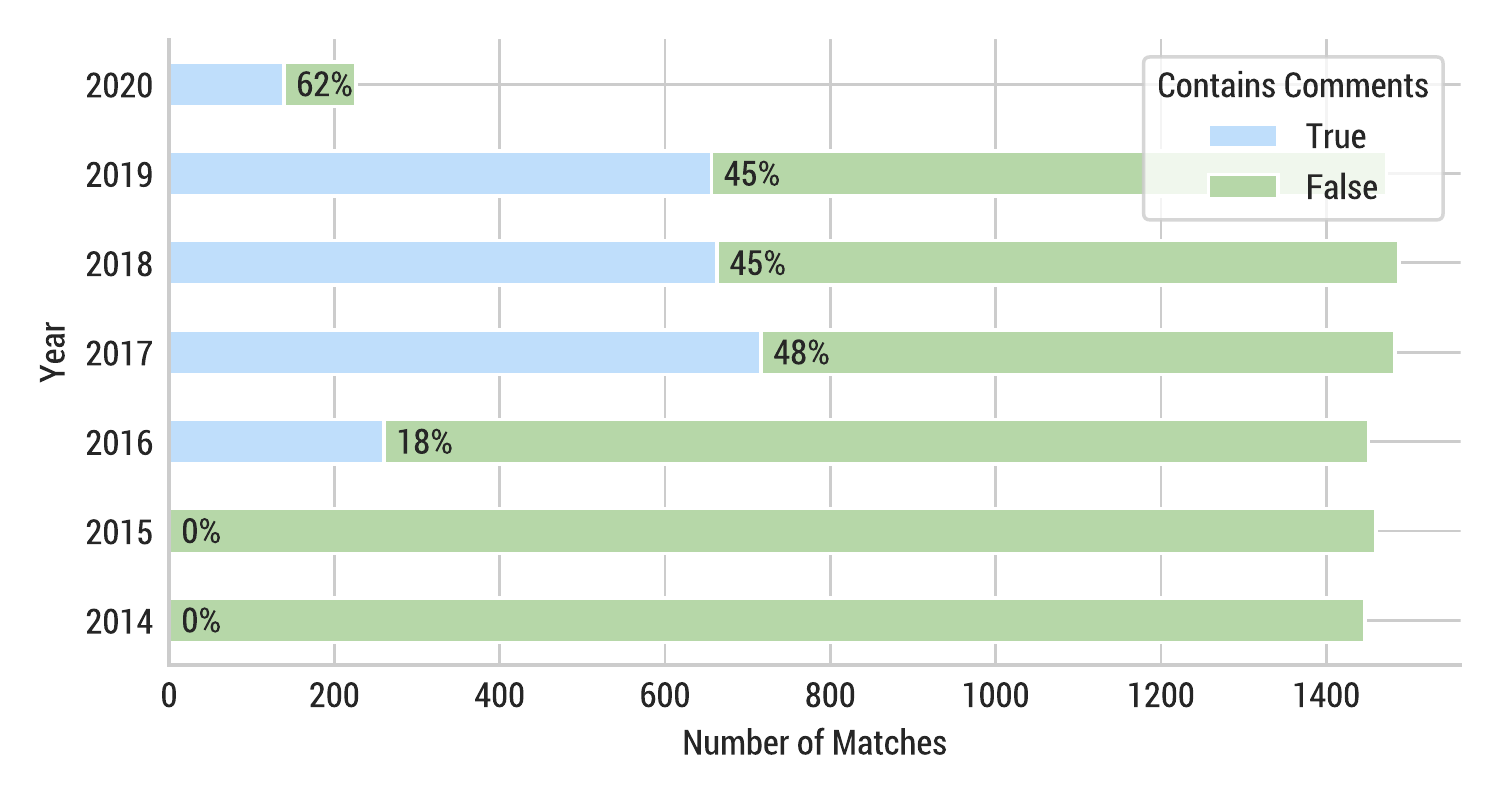}
    \caption{\label{fig:match_num} Frequency distribution of matches with and without commentary across available data years.}
    \vspace{-2mm}
\end{figure}

\section{Dataset Construction}
We collect time-stamped commentary with key events of 2,263 soccer matches in total. The matches stem from four major soccer tournaments including the UEFA Champions League, UEFA Europa League, Premier League and Series A between 2016 and 2020. \texttt{SOCCER} consists of over 135,000 time-stamped pieces of commentary and 31,000 within-match events. This section describes our data collection and preparation process in detail.

\begin{figure}[t]
\centering
    \includegraphics[width=1\linewidth]{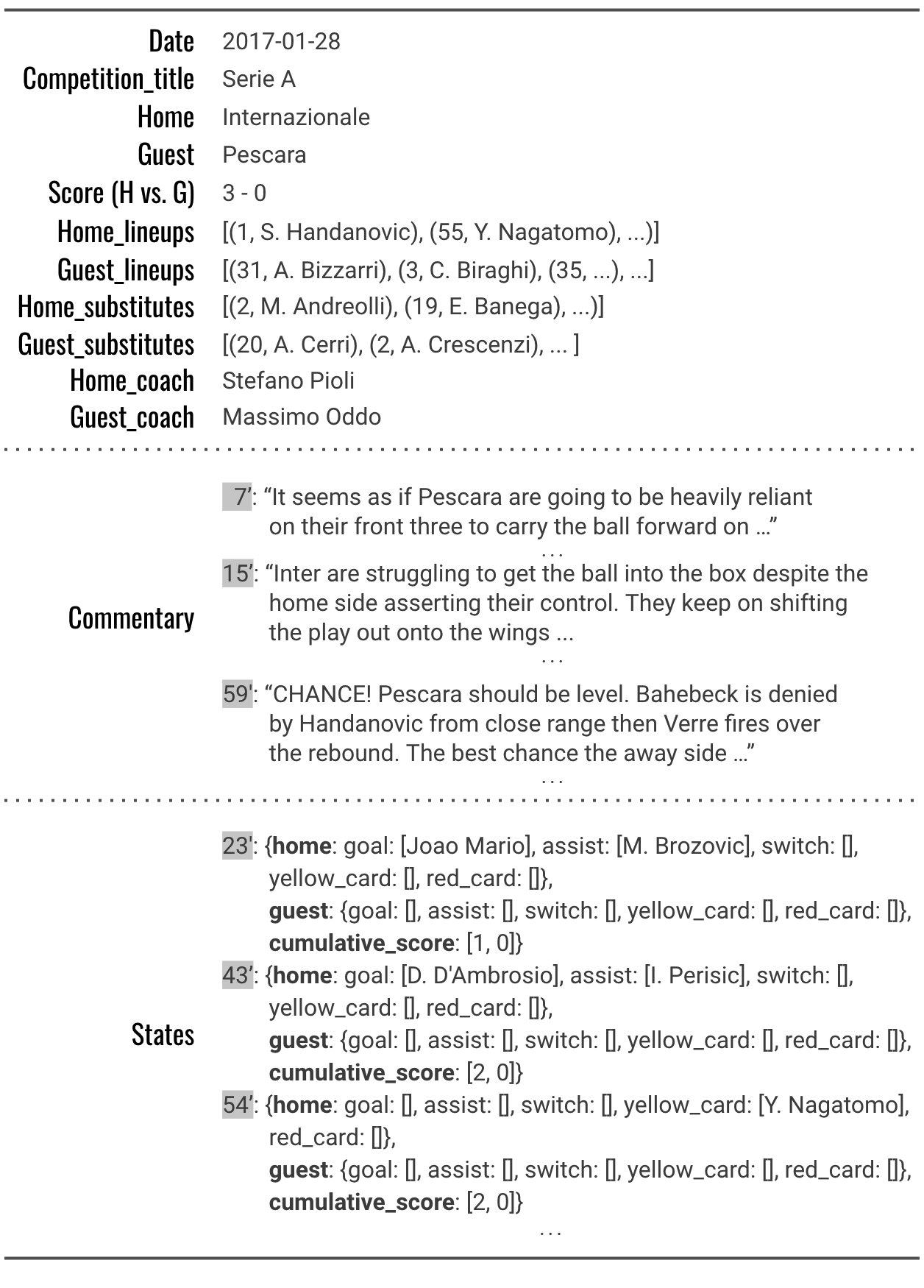}
    \caption{\label{fig:snippet} A short snippet of \protect\href{https://www.goal.com/en-us/match/internazionale-v-pescara/commentary-result/1gnkkzqjipx08k2ygwzelipyh}{a match} in the dataset. }
    \vspace{-2mm}
\end{figure}

\input{table}

\subsection{Data Processing}
Commentaries, events, team lineups, match dates and other meta-information are gathered from match-specific pages. Out of a total of 9,028 matches covered on \href{https://www.goal.com/en-us}{goal.com} between 2014 and 2020, we retain only those 2,434 matches that list detailed event records and commentary. Any matches missing either of the two information streams are discarded. The retained matches belong to the four major tournaments mentioned above and all occurred starting 2016. Figure~\ref{fig:match_num} shows the frequency distribution of included and overall matches across the years in which they took place. All commentaries are in English and available in text form, thus requiring no transcription. Pieces of commentary come pre-segmented and aligned to match-internal timestamps so that in-game events and commentary with the same timestamps can be linked. Comments whose temporal information is unavailable usually belong to the pre-game, intermission and post-game periods and are labeled as START, BREAK, END accordingly. The total number of commentary paragraphs within a game is the same as the number of timestamps. This number varies between matches as timestamps during which the commentator did not provide commentary are omitted. Finally, any templated sentences following the format ``\textit{team 1 score - score team 2}'' are removed to avoid trivial leakage of the match state. All annotation and filtering  processes are done programmatically and no manual efforts are involved.

Events are classified into five types: \textit{goal}, \textit{assist}, \textit{yellow card}, \textit{red card} and \textit{switch}. We consider events as keys and the event-related players as the corresponding values. For example, if player B from the home team assists in scoring a goal, player B will be the value of the event \textit{assist} for the home team. Hence, at each timestamp $t$, there are ten event-player pairs (five event types tracked for two teams). From this representation, we construct a comprehensive game state incorporating all the event-player pairs for each team as well as a cumulative score at each timestamp (See Figure~\ref{fig:snippet}). Special events such as penalty goals or own goals are not explicitly labeled, but can be derived from the evolution in cumulative score between neighboring timestamps. After processing, 171 games were found to have ill-formed commentary or mis-aligned end-game match scores compared to the goal records in the key events. These matches were eliminated from the original 2,434 games crawled with commentary, giving us a total of 2,263 games. Finally, the collected data is partitioned into distinct training (70\%), validation (15\%) and test (15\%) sets.

\begin{figure*}[t]
\centering
    \includegraphics[width=1\linewidth]{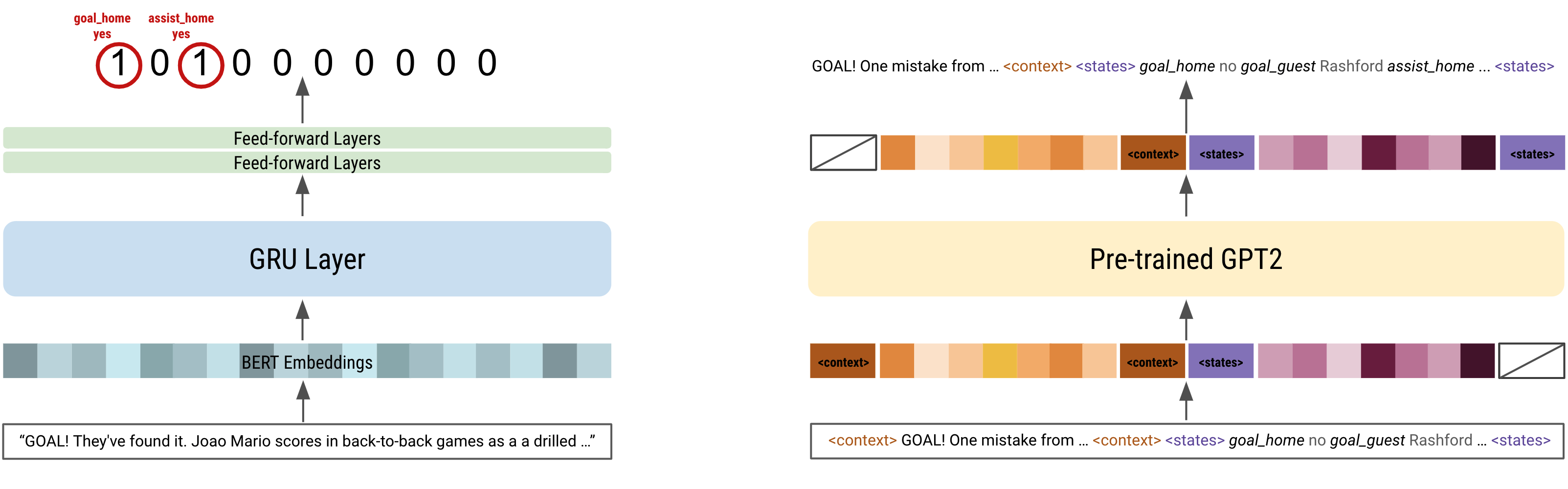}
    \caption{\label{fig:model} Model architecture of the GRU classifier and GPT-2 based variant.}
    \vspace{-3mm}
\end{figure*}

\section{State Definition and Task Proposal}
For each match $m$ in the dataset $M$, there is a set of timestamps $T_m = \{t\}$ accurate to a minute. As input, we are given a stream of commentaries $C_m = \{c_t\}_{t=1}^{T_m}$ and $c_t$ represents the paragraph of commentary at time $t$. The output will be a set of general match states $S_m = \{s_t\}_{t=1}^{T_m}$ such that each $s_t$ reflects the state change in the comment $c_t$ at the same timestamp. $s_t$ contains a set of events $e_{i, j}^{(t)}$, where $i$ represents the event types ($i \in \{\textit{goal},\textit{assist},\textit{yellow card}, \textit{red card}, \textit{switch}\}$) and $j$ denotes the event actor ($j \in \{\textit{home}, \textit{guest}\}$). Given the sparse distribution of $s_t$, we propose two alternative variants of the variable to assess the difficulty of state tracking at different granularity levels of state resolution.

\paragraph{Team Level.} In this simplest notion of state, events are tracked at the team level. In other words, $e_{i, j}^{(t)} = \{\textit{yes}, \textit{no} \}$. Consider the event of the home team scoring a goal $e_{\mathit{goal},\ \mathit  {home}}^{(t)}$ at time $t$ as an example: given the commentary $c_t$ and other related meta-information, a model is tasked with determining the value of $e_{\mathit{goal},\ \mathit{home}} ^{(t)}$ to be $\textit{yes}$ if the home team indeed scored a goal in a given minute, or $\textit{no}$ otherwise.

\paragraph{Player Level.} At this significantly increased level of resolution, all events are additionally associated with their player agents $p \in P$, where $P$ denotes the collection of players. Concretely, the variable $e_{i, j}^{(t)}$ is mapped to either the related players' names $p$ or a $none$ answer to each event at time $t$. To facilitate this form of state, match meta-information includes lineups that associate present players with teams.

\section{Analysis and Baseline Experiments}
In the following, we provide descriptive statistics of the \texttt{SOCCER} dataset and include two model baselines for recognizing match events resulting in changes of states.

\subsection{Dataset Statistics and Comparison}
\label{sec:sparse-info}
The \texttt{SOCCER} dataset covers 2,263 matches with 135,805 pieces of commentary and 31,542 in-game event records. In all event records, each event type of each team appears approximately 3,154 times on average. There are a total of 3,507 unique player names across all event types and an average 1,219 unique player names per event type per team. A more detailed overview of the distribution of event types and player names can be seen in Table~\ref{tab:data_stats}. 

Common state tracking datasets either in dialogue systems or procedural texts are designed to capture frequent state changes in the text. In \texttt{SOCCER}, we study a more general setting where the corpus is much less information dense due to an abundance of non-event related chatter. To quantify this difference, we define information density (\textit{ID}) as: 

\[ID = \frac{\textit{Total \# of state changes}}{\textit{Total \# of turns/steps/timestamps}}\]

As shown in Table~\ref{tab:info-dense}, our dataset has a considerably lower information density with more turns of information. In \texttt{SOCCER}, the match state only gets updated every 5 timestamps, while in datasets such as MultiWOZ2.1 \citep{eric2019multiwoz} and OpenPI \citep{tandon2020dataset}, there are between 1 and 4 state changes per turn or step on average.
\input{info_dense}
\subsection{Baseline Setup}
\texttt{SOCCER} presents a new challenge to the state tracking community by introducing a more general corpus with an all-new state definition and a sparse information distribution. These properties render it difficult to directly apply some existing models such as TRADE used in DST tasks and ProLocal \citep{dalvi-etal-2018-tracking} proposed for procedural texts. Motivated by previous work on state tracking and based on the characteristics of the task, we use two baseline training and inference schemes: 1) a GRU \citep{cho2014learning} classifier with pre-trained BERT \citep{devlin-etal-2019-bert} embeddings, and 2) a generative pre-trained GPT2 \citep{radford2019language} variant.

\textbf{\resizebox{0.85\linewidth}{!}{GRU Classifier with BERT Embeddings.}}
The GRU model is used as a preliminary baseline to assess the difficulty level of the \texttt{SOCCER} dataset. Embeddings of the timestamped commentary $c_t$ are obtained from the pretrained weights of BERT \citep{devlin-etal-2019-bert}, that then get fed into a 1-layer GRU \citep{cho2014learning} network followed by two feed-forward layers. We only tasked this model with team-level state tracking as the classification will be extremely difficult if each player name is treated as a distinct class. We map the 10 event variables $e_{i, j}^{(t)}$ as binary flags to a 10-bit scalar value in which each digit denotes the predicted value of a variable. For example, if the 0th position corresponds to the variable $e_{\mathit{goal},\ \mathit{home}}^{(t)}$, then the predicted value at that position denotes whether the home team scores a goal (See Figure~\ref{fig:model}). Compared to converting the problem into ten binary classifications, this allows us to directly model the joint occurrence of events.

\textbf{{GPT-2 Based Variant.}}
Recent approaches to state tracking \citep{kim2019efficient,hosseini2020simple,tandon2020dataset} have shown that generative models are competitive especially in open-vocabulary settings. Inspired by simpleTOD \citep{hosseini2020simple} and the OpenPI baseline \citep{tandon2020dataset}, we cast the player-level state tracking task as a sequence generation problem, allowing us to leverage the capabilities of causal language models such as GPT-2 \citep{radford2019language}. The training sequence consists of a concatenation of the commentary, event types and player names, allowing us to model the joint probability of the whole sequence. Event names are preprocessed as tokens like \textit{goal\_home} to avoid being tokenized into sub-word units. Commentary and event-player pairs are encapsulated in special tokens to help the model distinguish context from labels. See Figure~\ref{fig:model} for a schematic overview of the model training input. In training, the model takes the concatenated sequence as input to perform next token prediction task. At inference time, greedy decoding is used to generate state predictions due to its superior performance compared to beam search and top-k sampling \cite{hosseini2020simple}.

\subsection{Implementation Details}
During preprocessing, we find that 98.1\% of comments in the collection are shorter than 200 words, therefore any outliers with a length of more than 200 words are truncated at that point. Then, the input text sequences are tokenized using byte-pair encoding \citep{sennrich-etal-2016-neural} to avoid out-of-vocabulary words.

The sentence embeddings processed by the GRU classifier stem from the pretrained weights of HuggingFace's BERT model \citep{Wolf2019HuggingFacesTS}. The GPT-2 model \citep{radford2019language} is also obtained from HuggingFace with pretrained weights, which are then fine-tuned on \texttt{SOCCER}\footnote{The \texttt{SOCCER} dataset as well as the code base used to collect it and run the experiments presented in the remainder of this paper are available \href{https://github.com/bcbi-edu/p_eickhoff_SOCCER}{here}.}. 

\input{general_table}

\subsection{Evaluation}\label{eval}
Accuracy, and recall for occurrences of all event-types are used to assess the performance of both models. Due to the sparsity of event occurrences, recall is crucial to track the models' ability to extract events given the full set of types. For convenience, we refer to event types with ground truth $none$ answers as negative cases and positive cases otherwise. Therefore, recall among event occurrences is referred to as positive recall in the tables. More specifically, in Tables \ref{tab:all-scores} and \ref{tab:density}, accuracy and positive recall are measured on all labels (positive and negative combined). In Table \ref{tab:per-event}, the performance is reported on positive labels only, and detailed metrics including precision, recall and F1 scores are provided.
\input{per-event}
\input{densities}

\section{Results}
This section reports the results on the test set of \texttt{SOCCER}. As a na\"{i}ve distributional baseline, we compute the ratio of negative cases in the test set to be 0.9766. 

In Table~\ref{tab:all-scores}, both models achieve an accuracy that is approximately equal to this majority class baseline due to the heavily imbalanced distribution of event positives and negatives. While accuracy scores are very high, positive recall is much lower, indicating that many event occurrences are missed by the models. When comparing the GPT-2 model's performance on both team level and player level event recognition\footnote{The GRU classifier is only used in team-level tasks since treating each player in the ontology as a distinct class to classify is very difficult.}, we notice that player level recall is substantially worse than that on team-level. This result suggests that complex state tracking involving broad ranges of possible slot values is a comparatively harder task that may require more sophisticated approaches. 

\subsection{Results Per Event Type}
In addition to these general results, we break down model performance of positive cases by event-type and provide additional metrics including precision, recall and $F_1$ scores (see Table~\ref{tab:per-event}). When associating the scores with the event type distribution (see Table~\ref{tab:data_stats}), we can observe that, generally, greater numbers of available data points result in better performance. Take the event type \textit{goal} as an example. According to Table~\ref{tab:data_stats} there are about 800 more positive cases of the event  $e_{\mathit{goal},\ \mathit{home}} ^{(t)}$ than  $e_{\mathit{goal},\ \mathit{guest}} ^{(t)}$. A difference that is reflected in all the metrics in Table~\ref{tab:per-event} for both models. Another interesting point to note is the performance gap between the GRU classifier and GPT-2 model on the event type \textit{red card}. The \textit{red card} event is extremely rare in \texttt{SOCCER} as illustrated in Table~\ref{tab:data_stats}. Though we observe the performance of both models on \textit{red card} events to be comparably lower than those of the other events, the GRU classifier is able to capture more positive cases while no occurrences are detected by GPT-2.

\subsection{Results on Varying Information Densities}
In Section~\ref{sec:sparse-info}, we have shown that a key difference between \texttt{SOCCER} and other state tracking datasets lies in its low information density (See Table~\ref{tab:info-dense} for a detailed comparison). It is conceivable that such differences in information density affect state tracking performance. To eliminate confounding effects introduced via direct comparison to other datasets, this section explores the connection between event density across pieces of commentary and model performance. We begin by discarding all but the truly event related comments in each match to obtain a subset containing 0\% negative cases. This subset contains 25,934 event related comments across all matches. Then, by randomly replacing positive comments \footnote{Positive comments here refer to comments with event occurrences.} with negative ones from the same match at a sparsity ratio $r \in \{20\%, 40\%, 60\%, 80\%\}$, we keep the total number of comments at the same constant count of 25,934 and keep the temporal ordering of comments intact, while effectively reducing the level of information density. Table~\ref{tab:density} reports accuracy and positive recall for both methods and task levels when training and evaluating on non-overlapping splits of the newly constructed subsets. Note that, despite our earlier discussion of information density, Table~\ref{tab:density} reports a converse notion, sparsity. In this setting, 0\% corresponds to the highest and 80\% the lowest information density. 

Comparing accuracy at different event sparsity levels, we notice that scores increase as events become more sparsely distributed. This effect stems from the fact that, when we are replacing event related comments with non-event chatter, chance agreement improves as the number of true negatives increases. Positive recall of event occurrences, however, demonstrates an opposing trend, suggesting that the task of recognizing true state updates becomes more challenging the sparser the discourse domain is. This assumption is further supported by the different degree of performance observed on \texttt{SOCCER} vs.\ existing collections such as MultiWOZ2.1 \citep{eric2019multiwoz}, where recall scores of many models range in the mid-fifty percent range.

\section{Conclusion}
In this paper, we introduce \texttt{SOCCER}, the first discourse state tracking collection in the sports commentary domain. We propose two different levels of state granularity and provide two performance benchmarks for models ranging from GRU \citep{cho2014learning} for embedding temporal dependency to GPT-2 \citep{radford2019language} for causal language modeling. The dataset shows a much lower information density than many existing resources on state tracking, making it considerably more challenging. We believe that, in conjunction with the wide vocabulary of player-level notions of state, this property makes \texttt{SOCCER} an exciting resource on which our community can advance discourse state tracking to a broader range of settings than have been studied previously.

\section*{Acknowledgement}
This research is supported in part by the NSF (IIS-1956221). The views and conclusions contained herein are those of the authors and should not be interpreted as necessarily representing the official policies, either expressed or implied, of NSF or the U.S. Government. We would like to thank Ellie Pavlick, Stephen Bach, Zejiang Shen and the anonymous reviewers for their constructive feedback and helpful discussion.

\bibliographystyle{acl_natbib}
\bibliography{anthology,custom}
\end{document}

%% file: table.tex
\begin{table*}[h!]
  \centering
  \resizebox{1.\linewidth}{!}{
    \begin{tabular}{lcccccccccc}
    \hline
    \textbf{Event Name} &
      \multicolumn{2}{c}{\textbf{Goal}} &
      \multicolumn{2}{c}{\textbf{Assist}} &
      \multicolumn{2}{c}{\textbf{Yellow Card}} &
      \multicolumn{2}{c}{\textbf{Red  Card}} &
      \multicolumn{2}{c}{\textbf{Switch}} \\ \hline
    \textbf{Team} &
      \textit{Home} &
      \textit{Guest} &
      \textit{Home} &
      \textit{Guest} &
      \textit{Home} &
      \textit{Guest} &
      \textit{Home} &
      \textit{Guest} &
      \textit{Home} &
      \textit{Guest} \\ \hline
    \textbf{Event \# per team}  & 3582        & 2799       & 2434        & 1871       & 3948        & 4320       & 163        & 197        & 6111        & 6117        \\
    \textbf{Event total \#}     & \multicolumn{2}{c}{6381} & \multicolumn{2}{c}{4305} & \multicolumn{2}{c}{8268} & \multicolumn{2}{c}{360} & \multicolumn{2}{c}{12228} \\ \hline
    \textbf{Player \# per team} & 1001        & 924        & 882         & 774        & 1548        & 1613       & 145        & 183        & 2546        & 2575        \\
    \textbf{Player total \#}    & \multicolumn{2}{c}{2915} & \multicolumn{2}{c}{1656} & \multicolumn{2}{c}{3161} & \multicolumn{2}{c}{328} & \multicolumn{2}{c}{5121}  \\ \hline
    \end{tabular}
  }
\caption{\label{tab:data_stats} Event type and player name distribution.}
\vspace{-3mm}
\end{table*}

%% file: info_dense.tex
\begin{table}[]
    \centering
    \resizebox{1.\linewidth}{!}{
    \begin{tabular}{lccc}
    \hline
                          & \textbf{MultiWOZ2.1} & \textbf{OpenPI} & \textbf{SOCCER} \\ \hline
\textbf{Avg. turn length} & 11.75 - 15.12 & 13.32 & 50.28 \\
    \textbf{Avg. turn \#}  & 13.46             & 5.00             & 60.01           \\
    \textbf{\textit{ID}} & 1.05              & 3.8 - 4.3       & 0.19            \\ \hline
    \end{tabular}
    }
    \caption{\label{tab:info-dense} Information density of \texttt{SOCCER} vs.\ other state tracking datasets.}
    \vspace{-2mm}
\end{table}

%% file: general_table.tex
\begin{table}[]
\resizebox{1.\linewidth}{!}{
\begin{tabular}{lcccc}
\hline
                 & \multicolumn{2}{c}{\textbf{Team Level}} & \multicolumn{2}{c}{\textbf{Player Level}} \\ \hline
\textbf{Metrics} & \textit{Acc.}   & \textit{Pos. Recall}  & \textit{Acc.}    & \textit{Pos. Recall}   \\ \hline
\textbf{GRU classifier} & 0.9775 & 0.3990 & - & - \\ 
\textbf{GPT-2 variant}  & 0.9759 & 0.4855 & 0.9670  &  0.0775 \\
\hline
\end{tabular}
}
\caption{Team and player level test set performance.}
\label{tab:all-scores}
\vspace{-3mm}
\end{table}

%% file: per-event.tex
\begin{table*}[]
\centering
    \resizebox{1.\linewidth}{!}{
\begin{tabular}{llcccccccccc}
\hline
\textbf{} &
  \textbf{Event Type} &
  \multicolumn{2}{c}{\textbf{Goal}} &
  \multicolumn{2}{c}{\textbf{Assist}} &
  \multicolumn{2}{c}{\textbf{Yellow Card}} &
  \multicolumn{2}{c}{\textbf{Red Card}} &
  \multicolumn{2}{c}{\textbf{Switch}} \\ \hline
\textbf{Model} &
  \textbf{Team} &
  \textit{Home} &
  \textit{Guest} &
  \textit{Home} &
  \textit{Guest} &
  \textit{Home} &
  \textit{Guest} &
  \textit{Home} &
  \textit{Guest} &
  \textit{Home} &
  \textit{Guest} \\ \hline
\textbf{GRU Classifier}   & \textit{precision} & 0.54 & 0.50 & 0.51 & 0.33 & 0.46 & 0.54 & 0.13 & 0.19 & 0.55 & 0.53 \\
\textbf{}      & \textit{recall}    & 0.83 & 0.01 & 0.84 & 0.00 & 0.53 & 0.28 & 0.04 & 0.33 & 0.50 & 0.62 \\
\textbf{}      & \textit{$F_1$}        & 0.66 & 0.02 & 0.64 & 0.01 & 0.49 & 0.37 & 0.06 & 0.24 & 0.52 & 0.57 \\ \hline
\textbf{GPT-2 Variant} & \textit{precision} & 0.56 & 0.32 & 0.42 & 0.27 & 0.48 & 0.39 & 0.00 & 0.00 & 0.53 & 0.46 \\
\textbf{}      & \textit{recall}    & 0.89 & 0.08 & 0.92 & 0.06 & 0.95 & 0.06 & 0.00 & 0.00 & 0.93 & 0.01 \\
\textbf{}      & \textit{$F_1$}        & 0.69 & 0.13 & 0.58 & 0.09 & 0.64 & 0.11 & 0.00 & 0.00 & 0.67 & 0.01 \\ \hline
\end{tabular}
}
\caption{Model performance of positive classes on team-level task per event type.}
\label{tab:per-event}
\end{table*}

%% file: densities.tex
\begin{table*}[]
\resizebox{1.\linewidth}{!}{
\begin{tabular}{lcccccccccc}
\hline
\textbf{Comment Sparsity} &
  \multicolumn{2}{c}{\textbf{0\%}} &
  \multicolumn{2}{c}{\textbf{20\%}} &
  \multicolumn{2}{c}{\textbf{40\%}} &
  \multicolumn{2}{c}{\textbf{60\%}} &
  \multicolumn{2}{c}{\textbf{80\%}} \\ \hline
\textbf{Metrics} &
  \textit{Acc.} &
  \textit{Pos. Recall} &
  \textit{Acc.} &
  \textit{Pos. Recall} &
  \textit{Acc.} &
  \textit{Pos. Recall} &
  \textit{Acc.} &
  \textit{Pos. Recall} &
  \textit{Acc.} &
  \textit{Pos. Recall} \\ \hline
\textbf{Task Level}     & \multicolumn{10}{c}{\textbf{Team Level}}                            \\ \hline
\textbf{GRU Classifier} & 0.89 & 0.44 & 0.90 & 0.41 & 0.92 & 0.35 & 0.94 & 0.30 & 0.97 & 0.31 \\
\textbf{GPT-2 Variant}  & 0.88 & 0.49 & 0.90 & 0.49 & 0.93 & 0.47 & 0.95 & 0.41 & 0.98 & 0.44 \\ \hline
\textbf{Task Level}     & \multicolumn{10}{c}{\textbf{Player Level}}                          \\ \hline
\textbf{GPT-2 Variant}  & 0.83 & 0.06 & 0.87 & 0.06 & 0.90 & 0.04 & 0.94 & 0.04 & 0.98 & 0.02 \\ \hline
\end{tabular}
}
\caption{Model performance on team-level and player-level tasks with data of different information density.}
\label{tab:density}
\vspace{-3mm}
\end{table*}